# Deep Learning Approach for Predicting the Replicator Equation in Evolutionary Game Theory

Advait Chandorkar[1]

[1]Department of Mechanical Engineering, Indian Institute of Technology Ropar, Rupnagar, India
Advaitc07@gmail.com

## Abstract

*This paper presents a physics-informed deep learning approach for predicting the replicator equation, allowing accurate forecasting of population dynamics. This methodological innovation allows us to derive governing differential or difference equations for systems that lack explicit mathematical models. We used the SINDy model first introduced by Fasel, Kaiser, Kutz, Brunton, and Brunt 2016a to get the replicator equation, which will significantly advance our understanding of evolutionary biology, economic systems, and social dynamics. By refining predictive models across multiple disciplines, including ecology, social structures, and moral behaviours, our work offers new insights into the complex interplay of variables shaping evolutionary outcomes in dynamic systems*

## 1. Introduction

Game theory helps to understand how strategic behaviours evolve and persist in biological, social, and economic systems where individuals interact. It also helps in how complex social behaviours and strategies can evolve and persist in diverse contexts. Analysing in the principles of Nash equilibrium (NE) defined by John F. and Nash, Jr (1) offers insights into the stability and evolution of strategies within populations, shedding light on how cooperative or competitive behaviours evolve and endure across various contexts. This approach not only enriches theoretical understanding but also informs practical strategies in fields ranging from biology to economics.

The concept of evolutionary game theory was first proposed by John Maynard Smith and George R Martin (2) which extends classical game theory by focusing on the dynamics of strategy evolution within populations over time. Unlike traditional game theory, which typically analyses the strategic interactions of rational, decision-making individuals aiming to maximize their payoffs, EGT considers populations of interacting agents whose strategies evolve based on their success. In EGT, strategies represent phenotypic traits, such as aggression or body size, that impact fitness and survival. The success of a strategy is determined by its reproductive success in the population, influenced by selective pressures. This approach permits the mathematical analysis of the development of strategies, giving us an understanding of the evolutionary stability of behaviours and characteristics in biological populations. The replicator equation plays a key role in evolutionary game theory. used to model the dynamics of strategy distribution within a population. The replicator is a non-linear differential equation that allows us to capture the realworld strategies and interactions. The replicator dynamics can be used to find out how the percent of people using one particular strategy varies over time and other parameters like changing payoffs. Beyond its classical applications, the replicator equation has been employed in diverse optimization and control scenarios, such as bandwidth allocation (4) and service levels in water distribution systems (5).

Predicting the behaviour of non-linear differential equations from their trajectories is challenging due to non-linear systems often exhibit sensitivity to initial conditions, where small variations in starting points can lead to vastly different outcomes due to system the system being chaotic (butterfly effect). This sensitivity makes long-term prediction unreliable due to finite precision error. Third, the presence of multiple equilibrium and bifurcations in non-linear systems complicates the analysis, as trajectories can converge to different stable states depending on initial conditions and perturbations. Non linear systems can also have tipping points which by just monitoring the population we won't be able to exactly predict when the tipping point may occur.

A method discovered by Bongard and Lipson (5), as well as Schmidt and Lipson (6), presented a method for uncovering the underlying structure of nonlinear dynamical systems from data. This approach used genetic programming (7) to derive nonlinear differential equations. Additional methods for recognizing dynamic systems from data include methods for extracting governing equations from time-series data equation-free modelling (7), empirical dynamic modelling (9, 10), modelling emergent behaviours (11), and automated dynamic inference (11–13).

In this paper, the aim is to predict the replicator equation using the SINDy (Sparse Identification of Nonlinear Dynamical systems) architecture that can be used to solve ordinary and partial differential equations and validate its use in this field, first introduced by Steven L. Brunton, Joshua L. Proctor, and J. Nathan Kutz in (8). Sparse identification of nonlinear dynamical systems (SINDy) uses the fact that the sparsity of most physical systems, only have few functions as their basis function to define the dynamics which is generally the case in Evolutionary game theory, resulting in governing equations that are sparse in a high-dimensional nonlinear function space. The SINDy framework identifies these sparse representations efficiently, even in complex systems, and has demonstrated success in capturing the essential dynamics with minimal computational overhead. Additionally, a weak form of SINDy was developed by A.Messenger, Dall'Anese(14) Bortzto handle noisy data, as there is always some noise in the data collected from real world applications. SINDy can be applied to a variety of contexts, including discretetime systems, nonlinear systems, and other complex dynamical systems, making it a versatile tool for discovering underlying dynamics from data. Its adaptability to different types of systems and robustness against noise make it useful for finding the replicator equation.

## 2. NETWORK ARCHITECTURE

The sparse identification of nonlinear dynamical systems uses the fact that most differential equations have very few terms that define the dynamics, making the governing equations sparse in a high-dimensional nonlinear function space uses sparse regression techniques to reduce the unwanted terms in the differential equation due to noise in the data. Progress in compressed sensing and sparse regression has made the concept of sparsity advantageous. These advancements allow us to identify non-zero terms without resorting to a combinatorial, intractable brute-force search. This ensures that we can find the sparse solution with high probability using convex methods that efficiently scale to large problems, in line with Moore's law. Consequently, the identified sparse model strikes a balance between complexity (i.e., the sparsity of the righthand-side dynamics) and accuracy, thus preventing overfitting to the data.

## 2.1. Data

The sparse identification of nonlinear dynamical systems takes either *x(t)* or *x˙(t)* for continuous time, or *x[n]* or Δ*x[n]* for discrete-time. The sparse identification of nonlinear dynamical systems takes either *x(t)* or *x˙(t)* for continuous time, or *x[n]* or Δ*x[n]* for discrete-time. The *x* and *x˙* sampled over time are arranged in a matrix as shown below, where $t_1, t_2, ..., t_m$ are the time points sampled in increasing order.

$$\mathbf{X} = \begin{bmatrix} x^T(t_1) \\ x^T(t_2) \\ \vdots \\ x^T(t_m) \end{bmatrix} = \begin{bmatrix} x_1(t_1) & x_2(t_1) & \cdots & x_n(t_1) \\ x_1(t_2) & x_2(t_2) & \cdots & x_n(t_2) \\ \vdots & \vdots & \ddots & \vdots \\ x_1(t_m) & x_2(t_m) & \cdots & x_n(t_m) \end{bmatrix}$$

$$\dot{\mathbf{X}} = \begin{bmatrix} \dot{x}^T(t_1) \\ \dot{x}^T(t_2) \\ \vdots \\ \dot{x}^T(t_m) \end{bmatrix} = \begin{bmatrix} \dot{x}_1(t_1) & \dot{x}_2(t_1) & \cdots & \dot{x}_n(t_1) \\ \dot{x}_1(t_2) & \dot{x}_2(t_2) & \cdots & \dot{x}_n(t_2) \\ \vdots & \vdots & \ddots & \vdots \\ \dot{x}_1(t_m) & \dot{x}_2(t_m) & \cdots & \dot{x}_n(t_m) \end{bmatrix}$$

Then a matrix Θ(**X**) consisting of the functions that may be present in the basis functions **X**. needs to be constructed based on the functions that may be present in the differential equations

$$\Theta(X) = [1 \quad X^2 \cdots \sin(X) \cos(X)]$$

Its capabilities were extended to models trained using multiple dynamic trajectories, and the generation of many models with subsampling and ensembling methods by U. Fasel, J. N. Kutz, B. W. Brunton and S. L. Brunton(15) which will be useful when the dynamics of the replicator equation gets complicated helping it in accurately predicting the equation. One recent development is that researchers have started considering feedback like Joshua S. Weitza,b,1, Ceyhun Eksin a,c , Keith Paarpornc , Sam P. Brown, and William C. Ratcliff

## 3. METHODOLOGY

### 3.1. Data Generation

For training the SINDy model, we generated data by simulating the game we are testing on. Knowing the dynamics of the game, we derived the replicator equation to model the strategy evolution. To obtain a variety of trajectories, we employed a technique for generating random trajectories using Runge-Kutta algorithm. Then we converted to barycentric coordinates, which are particularly suited for capturing the proportions of strategies in a simplex.

### 3.2. Model Training and Prediction

In the context of replicator dynamics for n strategies, the sum of the strategy frequencies xi must always equal 1, i.e

$$\sum_{i=1}^{n} x_i = 1$$

This constraint allows us to reduce the number of equations that need to be predicted using the SINDy model. Instead of predicting all n equations, we only need to predict n-1 equations, as the final one can be determined directly from the constraint. There is no need to explicitly substitute one variable (example $x_n = 1 - \sum_{i=1}^{n-1} x_i$ ) into the system to match the solution for verification.

By reducing the number of equations, the process becomes computationally less taxing, especially for smaller models. This allows the already-built SINDy model to perform faster, without sacrificing accuracy, making it more efficient while maintaining the ability to accurately capture the system's dynamics.

## 4. REPLICATOR EQUATION

The replicator equation tells how individuals or populations change over time. The most famous game is of rock paper scissors with the origin of RPS has been difficult to trace, but there is some written evidence suggesting the Chinese played it already in the Han Dynasty more than 2000 years ago. There are three possible action choices: Rock(R), Paper (P), and Scissors (S) with the payoff given below in the table (1).

|   | R  | P  | S  |
|---|----|----|----|
| R | 0  | -1 | 1  |
| P | 1  | 0  | -1 |
| S | -1 | 1  | 0  |

The above payoff matrix gives a mixed strategy Nash equilibrium of (1/3, 1/3, 1/3) This simple game has many real-life examples. Colour polymorphism of male side-blotched lizards (17) is one of the real-world examples. Another instance is when European honeybees after being put in the local habitat of Japanese hornets in Japan, invaded the local honeybees but were not developed for attacks from Japanese hornets. In contrast, Japanese honeybees have developed a collective thermal defence mechanism against the hornets as a result of evolutionary adaptation from being in the same environmental place (18). The replicator equations for the Rock-PaperScissors game are given by:

$$\dot{x_R} = x_R(-x_P + x_S - f)$$

$$\dot{x_P} = x_P(x_R - x_S - f)$$

$$\dot{x_S} = x_S(-x_R + x_P - f)$$

where the finesses and the average fitness are:

It has evolved from symmetric games where all the individuals have the same payoff matrix and the same strategies available to non-symmetric games where the players have different strategies or payoffs assigned to them. One of the most popular non-symmetric games is the Battle of sexes. This game illustrates a coordination problem between two players with different preferences. The game involves two players who want to spend the evening together but have different preferences over two activities: football and ballet or any other conflicting hobbies. The husband prefers football, while the wife prefers ballet. Despite their different preferences, both players would prefer to be together than apart. This creates a situation where they need to coordinate their choices, leading to multiple equilibria.

|          | Football | Ballet |
|----------|----------|--------|
| Football | (2, 1)   | (0, 0) |
| Ballet   | (0, 0)   | (1, 2) |

In this matrix, the first number in each pair represents the payoff for Player 1 (the husband) and the second number represents the payoff for Player 2 (the wife). There are two pure-strategy Nash equilibria: (Football, Football) and (Ballet, Ballet) one mixed-strategy equilibrium where both players try to get the best outcome for themselves.

The following equations describe the replicator dynamics for this game. Let x represent the likelihood that Player 1 will select football, and y represent the likelihood that Player 2 will select football. The replicator equations for the strategies are:

$$\dot{x} = x(2y - x(2y + (1-x)(1-y)))$$

$$\dot{y} = y(x - y(1x + 2(1-y)(1-x)))$$

## 4. RESULTS AND DISCUSSION

### 4.1. RPS Game

In the Rock-Paper-Scissors (RPS) game, a Nash equilibrium (NE) occurs when players choose Rock, Paper, and Scissors with equal with equal probability to win, tie, or lose. While no player benefits from changing their strategy, the equilibrium's stability depends on the relative payoffs for winning, losing, and tying. If a strategy performs better than the population average, its frequency rises, while less successful strategies decline. This leads to cyclical dominance, where one strategy temporarily prevails but is eventually overtaken, causing the population to oscillate in a closed loop as shown by the simulation as show in Figure (1) below.

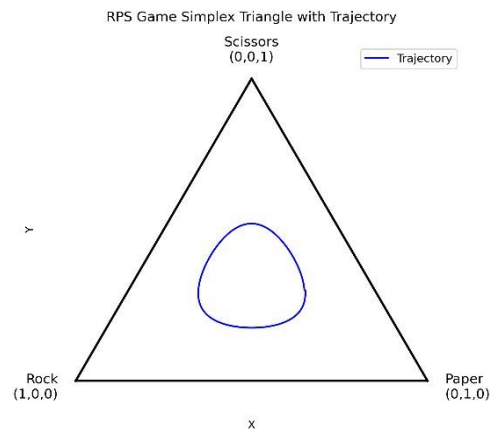

Fig. 1: RPS Game Simplex Triangle with Trajectory showing cyclical dominance

The oscillatory behaviour observed in the simulation reflects the dynamic interplay between the three strategies. Each strategy dominates for a short time but is ultimately replaced by another, ensuring that no single strategy is stable in the long run. This cyclical dominance is a key characteristic of many biological and social systems modelled by evolutionary game theory, where species, strategies, or behaviours rise and fall over time due to competition and adaptation.

For the RPS game, a single trajectory of 1000 data points was enough for sindy to predict the replicator equation SINDy is easily able to predict the replicator equation for RPS game taking a smaller number of data points than these correctly predict the basis functions but the coefficients were slightly off.

### 4.2. Battle of Sexe

The Battle of the Sexes game has two pure strategy Nash equilibria, where both players coordinate on the same outcome: Both players choose Outcome A. The payoffs are (2, 1). Both players choose Outcome B. The payoffs are (1, 2).

In addition to these pure strategy equilibria, there is a mixed strategy Nash equilibrium where both players play a mix of strategies. In this equilibrium, each player assigns probabilities to their choices such that they are indifferent to the other player's strategy. The mixed strategy equilibrium can be found by solving the following system:

Let p represent the likelihood that Player 1 choose Outcome A, and 1-p represent the likelihood that Player 1 selects Outcome B. Let q represent the likelihood that Player 2 choose Outcome A, and let 1-q represent the likelihood that Player 2 selects Outcome B. The mixed strategy equilibrium is given by:

$$P=1/3 \quad Q=2/3$$

In this equilibrium, each player chooses according to these probabilities, leading to expected payoffs where neither player has an incentive to deviate from their mixed strategy.

To analyse the dynamics of the Battle of the Sexes using replicator dynamics, we consider the evolution of the strategies over time. The replicator equation for this game describes how the proportion of players using each strategy changes based on their relative fitness.

For the battle of sexes game, the number of data points required were considerably more. In order to reduce the number of data points, multiple shorter trajectories were generated which gave better results compared to a single larger trajectory.

### 4. CONCLUSION

In conclusion, by applying the SINDy (Sparse Identification of Nonlinear Dynamics) framework, we can derive the replicator equation directly from observed data points (changes in the number of species). This approach enables us to get the underlying differential equation which will help us predicting the future outcomes.

By understanding how the population of a species evolves over time, it will become possible to identify critical tipping points where the population of one species may blow up or go extinct due to over-harvesting or environmental changes. The governing equation will provide insights for preventing a species from being driven to extinction in a local environment.

### DATA AVAILABILITY STATEMENT

The datasets and the code generated during and analysed during the current study are available from the corresponding author on reasonable request

### REFERENCES

[1] W. H. Thompson and R. A. Birney, "The Logical Form of the Law of Change," Proc. Nat. Acad. Sci., vol. 36, no. 1, pp. 48–54, Jan. 1950. doi: 10.1073/pnas.36.1.48.

[2] J. Maynard Smith and G. R. Price, "The logic of animal conflict," Nature, vol. 246, no. 5427, pp. 15–18, 1973.

[3] R. Cressman and Y. Tao, "The Replicator Equation and Other Game Dynamics," Proc. Nat. Acad. Sci., vol. 111, pp. 10810–10817, 2014. doi: 10.1073/pnas.23800665.

[4] J. Poveda and N. Quijano, "Dynamic bandwidth allocation in wireless networks using a shahshahani gradient based extremum seeking control," in Proc. 2012 6th Int. Conf. on Network Games, Control and Optimization (NetGCooP), Paris, France, 2012, pp. 44–50.

[5] E. Ramírez-Llanos and N. Quijano, "A population dynamics approach for the water distribution problem," Int. J. Control, vol. 83, no. 9, pp. 1947–1964, 2010.

[6] J. R. Koza, Genetic Programming: On the Programming of Computers by Means of Natural Selection, vol. 1. Cambridge, MA: MIT Press, 1992.

[7] J. P. Crutchfield and B. S. McNamara, "Equations of motion from a data series," Complex Syst., vol. 1, no. 3, pp. 417–452, 1987.

[8] I. G. Kevrekidis, C. W. Gear, and G. Hummer, "Equation-free, coarse-grained multiscale computation: Enabling microscopic simulators to perform system-level analysis," Commun. Math. Sci., vol. 1, no. 4, pp. 715–762, 2003.

[9] W. H. Thompson and R. A. Birney, "The Logical Form of the Law of Change," Proc. Nat. Acad. Sci., vol. 36, no. 1, pp. 48–54, Jan. 1950. doi: 10.1073/pnas.36.1.48.

[10] M. D. Schmidt, C. A. Myers, and H. Lipson, "Automated refinement and inference of analytical models for metabolic networks," Phys. Biol., vol. 8, no. 5, p. 055011, 2011.

[11] B. C. Daniels and I. Nemenman, "Automated adaptive inference of phenomenological dynamical models," Nat. Commun., vol. 6, p. 8133, 2015.

[12] B. C. Daniels and I. Nemenman, "Efficient inference of parsimonious phenomenological models of cellular dynamics using S-systems and alternating regression," PLoS One, vol. 10, no. 3, p. e011982, 2015.

[13] D. A. Messenger, E. Dall'Anese, and D. Bortz, "Online weak-form sparse identification of partial differential equations," in Proc. Math. Sci. Machine Learning, vol. 190, pp. 241–256, 2022

[14] J. Burton, R. Brockett, and B. S. Meyers, "Ensemble-SINDy: Robust sparse model discovery in the low-data, high-noise limit, with active learning and control," arXiv preprint arXiv:2111.10992, 2021

[15] H. J. Zhou, "The rock–paper–scissors game," Contemp. Phys., vol. 57, no. 2, pp. 151–163, 2015. doi: 10.1080/00107514.2015.1026556.

[16] B. Sinervo and C. Lively, "The rock-paper-scissors game and the evolution of alternative male strategies," Nature, vol. 380, pp. 240–243, 1996.

[17] M. Ono, T. Igarashi, E. Ohno, and M. Sasaki, "Unusual thermal defence by a honeybee against mass attack by hornets," Nature, vol. 377, 1995.

[18] A. R. Tilman, J. B. Plotkin, and E. Akçay, "Evolutionary games with environmental feedbacks," Nat. Commun., vol. 11, no. 915, 2020. doi: 10.1038/s41467-020-14531-6.

**Authors**

Advait Chandorkar

Advait Chandorkar is a fourth-year B.Tech. student at IIT Ropar, pursuing a degree in Mechanical Engineering.

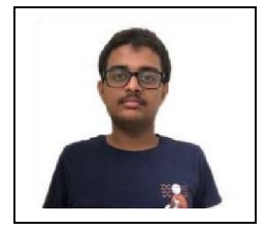